\documentclass[opre, nonblindrev]{style/informs} 

\OneAndAHalfSpacedXI 

\usepackage{graphicx}
\usepackage{diagbox}
\usepackage{subfigure, epsfig}
\usepackage{natbib}
\usepackage{xcolor}

 \bibpunct[, ]{(}{)}{,}{a}{}{,}%
 \def\bibfont{\small}%
\TheoremsNumberedThrough     
\ECRepeatTheorems

\EquationsNumberedThrough    

\usepackage{enumerate}
\usepackage{graphicx}
\usepackage{multirow}
\usepackage{color}
\usepackage{amsmath}
\usepackage{amsfonts}
\usepackage{amssymb}
\usepackage{mathrsfs}
\usepackage{mathptmx}
\usepackage{eucal,bm}
\usepackage{algorithm}

\usepackage{algorithmic}

\newcommand{\Oscr}{\ensuremath{\mathcal O}}

\newcommand{\E}{\ensuremath{\mathbb E}}
\newcommand{\EE}{\ensuremath{\mathbb E}}
\newcommand{\PP}{\ensuremath{\mathbb P}}

\newcommand{\cL}{\mathcal{L}}
\newcommand{\cT}{\mathcal{T}}

\newcommand{\ind}{\ensuremath{\mathbb I}}

\def\cO{{\cal O}}

\newcommand{\beq}{\begin{equation}}
\newcommand{\eeq}{\end{equation}}

\newcommand{\beqnr}{\begin{eqnarray}}
\newcommand{\eeqnr}{\end{eqnarray}}

\newcommand{\benum}{\begin{enumerate}}
\newcommand{\eenum}{\end{enumerate}}

\begin{document}


\RUNAUTHOR{Doe and Doe}

\RUNTITLE{POM Template}

\TITLE{Bridging Adversarial and Nonstationary  \\
Multi-armed Bandit}

\ARTICLEAUTHORS{%
 \AUTHOR{Ningyuan Chen}
 \AFF{Rotman School of Management, University of Toronto, Toronto, ON, Canada, \EMAIL{ningyuan.chen@utoronto.ca}} 
 \AUTHOR{Shuoguang Yang}
 \AFF{Department of Industrial Engineering and Decision Analytics, The Hong Kong University of Science and Technology,\EMAIL{yangsg@ust.hk}}
 \AUTHOR{Hailun Zhang}
 \AFF{School of Data Science, The Chinese University of Hong Kong, Shenzhen,  \EMAIL{zhanghailun@cuhk.edu.cn}}
} 

\ABSTRACT{%
In the multi-armed bandit framework, there are two formulations that are commonly employed to handle time-varying reward distributions: adversarial bandit and nonstationary bandit.  Although their oracles, algorithms, and regret analysis differ significantly, we provide a unified formulation in this paper that smoothly bridges the two as special cases.  The formulation uses an oracle that takes the best action sequences within a switch budget.  Depending on the switch budget, it turns into the oracle in hindsight in the adversarial bandit and dynamic oracle in the nonstationary bandit.  We provide algorithms that attain the optimal regret with the matching lower bound.  The optimal regret displays distinct behavior in two regimes. 
}%


\maketitle

%


\section{Introduction}\label{sec:intro}
The framework of online learning or multi-armed bandit is developed to handle the situation where the agent has to interact repeatedly with an unknown environment.
The agent has to simultaneously learn the rewards of available decisions (or arms) and search for the optimal arm.
The framework has seen great success in many applications, including dynamic pricing, clinical trials, and inventory management \cite{bubeck2012regret,den2015dynamic,bouneffouf2019survey}.

The classic multi-armed bandit considers the stationary environment,
in which the reward distribution of the arms remains constant over time.
This is hardly true in many applications.
For example, in dynamic pricing, the market environment that determines the reward of each arm (charged price) evolves constantly.
It causes the reward of each arm to drift as well.
To address this issue, there are two major frameworks proposed in the literature that do not impose any additional structures on the non-stationarity.
\begin{itemize}
    \item Adversarial bandit. The environment evolves in an arbitrary and adversarial way.
        As a result, the agent has to prepare for the worst-case realization of the mean reward over the horizon.
        To measure the performance of a policy, the \emph{regret} is calculated against a \emph{static oracle}, i.e., the best arm in hindsight over the whole period.
        The well-known EXP3 algorithm proposed in \cite{auer1995gambling} achieves the near-optimal regret $\tilde O(\sqrt{T})$\footnote{We use $\tilde O(\cdot)$ to denote the asymptotic rate omitting the logarithmic factors.}, where $T$ is the length of the learning horizon.
    \item Nonstationary bandit. The drift of the environment is controlled by a variational budget $V_T$.
        Note that $V_T=0$ corresponds to the stationary setting and $V_T=\Omega(T)$ corresponds to the adversarial setting.
        The policy needs to adapt to $V_T$ and its regret is measured against the \emph{dynamic oracle}, i.e., the best arm in each period.
        It has been shown in \cite{besbes2015non,besbes2019optimal} that the optimal regret is $\tilde O(V^{1/3}_TT^{2/3})$.
\end{itemize}
Comparing the two frameworks, there are two major differences.
First, in terms of the nonstationarity in the environment, the adversarial framework allows for more drastic changes.
In particular, the total variation over time can be linear in $T$.
On the other hand, the nonstationary bandit allows for a variational budget $V_T$ which must be sublinear in $T$; otherwise, it leads to linear regret from the above discussion.
As a result, the adversarial bandit is harder to learn.
Second, despite the difficulty, one can achieve smaller regret ($T^{1/2}$ versus $V_T^{1/3}T^{2/3}$) in the adversarial bandit, because the oracle against which the regret is measured is weaker.
The static oracle is the best single arm over the horizon in hindsight, unlike the dynamic oracle, which pulls the best arm in each period.
Therefore, the learning in the adversarial bandit has a lower standard than that in the nonstationary bandit.
The two differences drive distinct algorithmic designs, analyses, and regret bounds in the two formulations.

\subsection{Our Results}
Motivated by the difference of the oracles in the two formulations,
we propose a unified oracle that sits in between the static and dynamic oracles.
That is, we define an oracle that depends on ``switch" budget $S_T$, i.e., the optimal action sequences that will not switch more than $S_T$ times.
Specifically, the oracle takes the following form
\begin{equation*}
     \max\limits_{H(j_1,\cdots,j_t)\le S_T} \sum^T_{t=1}\mu_{t,j_t},
\end{equation*}
where $H(\cdot)$ is the hardness of the sequence $(j_1,\cdots,j_t)$, i.e., the number of switches throughout the sequence. 
It is clear that if $S_T= 1$, i.e., no switch is allowed, then this oracle reduces to the static oracle used in adversarial bandit;
if $S_T\ge T$, then it reduces to the dynamic oracle used in nonstationary bandit.

The switch budget $S_T$ has a natural interpretation in practice.
It essentially reflects the expectation the agent has for the algorithm.
For example, suppose the agent is a firm entering a new business and faces unknown market demand as well as seasonal
shifts in the market environment.
The arms are the prices the firm can charge and the reward is the profit.
If the firm is ambitious, then s/he may choose the switch budget to be the number of days a month and benchmark against the daily optimal prices.
Otherwise the firm may choose the switch budget to be one and benchmark against the monthly optimal prices.
Therefore, the choice of the switch budget reflects the conservativeness of the agent.
As a result, the regret shows how well the algorithm is measured by different degrees of conservativeness.

We design an algorithm for a switch budget $S_T$ and a variational budget $V_T$.
The algorithm attains the following regret.
When $S_T$ is large, i.e., $S_T\ge K^{-1/3} V_T^{2/3}T^{1/3}$ ($K$ is the number of arms), then the regret is bounded by $\cO ((KV_T)^{1/3}T^{2/3})$;
When $S_T$ is relatively small, i.e., $S_T\le  K^{-1/3} V_T^{2/3}T^{1/3}$, then the regret is bounded by $\cO (\sqrt{KS_TT})$.
The regret is shown to be near-optimal in the two regimes.

Our result implies that the regret has a phase transition.
When the switch budget is large, the regret doesn't depend on the switch budget and the regret is similar to that of nonstationary bandit;
when the switch budget is small, the regret doesn't depend on the variational budget and the regret is similar to that of adversarial bandit.
Table~\ref{tab:summary} summarizes four different cases according to the magnitudes of $S_T$ and $V_T$.
\begin{table}[htbp!]
    \centering
    \begin{tabular}{|l|l|l|} \hline
\backslashbox{$S_T$}{$V_T$} & $o(T)$ & $\Omega(T)$ \\ \hline
\multirow{2}{*}{$o(T)$}  &
$S_T< K^{-1/3} V_T^{2/3}T^{1/3}$: optimal regret $\cO(\sqrt{TS_T}) $ & \multirow{2}{*}{Adversarial bandit with regret $O(\sqrt{T})$} \\ 
 & $S_T\ge K^{-1/3} V_T^{2/3}T^{1/3}$: optimal regret $\cO(V_T^{1/3} T^{2/3})$ &  \\ \hline
$\Omega(T)$ & Non-stationary bandit  with regret $O(V_T^{1/3} T^{2/3})$ & No sublinear regret \\ \hline 
    \end{tabular}
    \caption{Regret analysis for different scenarios}
    \label{tab:summary}
\end{table}

\subsection{Related Literature}
This paper is related to the nonstationary bandit literature.
It extends the classic stochastic multi-armed bandit problems by introducing time-varying reward distributions of the arms.
Earlier papers consider discrete changes or a finite set of change points, such as \cite{auer2002nonstochastic, Garivier2011}, and thus is sometimes referred to as switching bandit.
Recent literature of this stream includes \cite{auer2019adaptively,chen2019new}.
In \cite{besbes2015non,besbes2019optimal}, the changes of the mean rewards can be continuous and a variational budget is imposed.
This setting is further investigated in various extensions, including continuous arms \cite{besbes2015stopt}, unknown variational budget \cite{cheung2019learning}, MDPs \cite{cheung2019non} and contextual bandit \cite{luo2018efficient}.
The oracle we consider in this work encapsulates the nonstationary bandit formulation considered in \cite{besbes2015non} as a special case.
Our novelty is mainly in the formulation of the unified oracle instead of the algorithmic design and analysis.
There are other formulations to incorporate the nonstationarity, such as \cite{wu2022performance,zhou2020regime}, which are not considered in this paper.

The framework of adversarial bandit is considered in \cite{auer1995gambling} and has been studied extensively in the last few decades.
\cite{bubeck2012regret} provides a review of the literature.
Our study is related to the literature on the design of algorithms that are agnostic to the faced bandit problem (adversarial or stochastic)
and can yet achieve optimal or near-optimal regret in both adversarial and stochastic bandit problems \cite{bubeck2012best,seldin2014one,auer2016algorithm}.
Our work differs from this literature in that the agent may face a continuous spectrum of formulations between the adversarial and stochastic (nonstationary) bandit problems.
The agent doesn't need to detect the specific formulation (associated with the sizes of the windows), which is given as prior information.
This paper is also related to the papers on stochastic bandit with adversarial corruption \cite{lykouris2018stochastic,gupta2019better,golrezaei2021learning}.
While this literature uses the dynamic oracle like the nonstationary bandit, we focus on the impact of the conservativeness of the oracle (switch budget) instead of adversarial changes in the distribution.

\section{Problem Formulation}\label{sec:formulation}
Let $[K] \triangleq \{1,2,\dots ,K\}$ be the set of arms and $T$ be the length of the horizon.
In each period $t\in [T]\triangleq \left\{1,\dots,T\right\}$, the agent chooses an action $A_t\in [K]$ and a random reward $Y_{t, A_t}$ is observed and collected.
Conditional on $A_t=k$, the random reward $Y_{t,k}\in [0,1]$ in period $t$ is independent of everything else and has mean $\mu_{t,k}\in[0,1]$, without loss of generality.

The agent does not know $\{\mu_{t,k}\}_{t\in[T],k\in[K]}$ initially.
She implements a policy $\pi$ to guide the decision in each period.
That is, $A_t = \pi_t(A_1,Y_{1,A_1},A_2,Y_{2,A_2}\dots,A_{t-1},Y_{t-1,A_{t-1}})$ is a mapping from the history prior to period $t$ to an arm in period $t$.
The objective of the agent is to maximize the total reward over the horizon $\sum_{t=1}^T Y_{t,A_t}$.

In the classic multi-armed bandit (MAB) framework, the expected total reward is benchmarked against the ``expected reward of the best arm'' when $\{\mu_{t,k}\}_{t\in[T],k\in[K]}$ were known.
Since we investigate the nonstationary case in which $\mu_{t,k}$ may vary over time, there are typically two ways to define the reward of the best arm(s).
One is the dynamic oracle $\sum_{t=1}^T \max_{k\in [K]}\mu_{t,k}$.
For the dynamic oracle, it pulls the best arm at time $t$:
\begin{equation*}
a_{t}^*= \argmax_{k \in [K]} \mu_{t,k}.
\end{equation*}
The other is the best arm in hindsight or a static oracle $\max_{k\in [K]}\sum_{t=1}^T \mu_{t,k}$.
Clearly, the dynamic oracle is a higher standard than the static oracle, which is targeted at the best arm in each period.
If $\mu_{t,k}$ varies arbitrarily, then without the knowledge, the agent cannot earn nearly as much reward as the dynamic oracle.
On the other hand, it has been shown in the adversarial bandit literature that algorithms can be developed to nearly match the static oracle in terms of regret.

In this study, we provide a framework to smoothly bridge the two types of oracles.
For a given (deterministic) arm sequence $a_1,\dots,a_T$, we define 
\begin{equation*}
    H(a_1,\dots,a_T) \triangleq \sum_{t=1}^{T-1} \ind(a_t\neq a_{t+1})+1.
\end{equation*}
In other words, $H(\cdot)$ measures the number of switches (plus one for convenience) in the arm sequence.
We define the \emph{unified oracle} as 
\begin{equation*}
  \Oscr(s,T, \{\mu_{t,k}\}_{t\in[T],k\in[K]}) \triangleq \max\{\sum^T_{t=1}\mu_{t,a_t}:H(a_1,\dots,a_T)\le s\}.
\end{equation*}
Note that $\Oscr(\cdot)$ is a deterministic function. It measures the performance of an arm sequence under the constraint that the number of switches in the sequence is at most $s-1$. 
The value of $s$ is the ``switch'' budget of the unified oracle.
\begin{remark}
	Our definition of unified oracle is inspired by the hardness function $H(\cdot)$ from \cite{auer2002nonstochastic}. To illustrate how to compute the unified oracle, consider the following simple example.
	Let $T= K= 3$, and $(\mu_{1,1}, \mu_{1,2}, \mu_{1,3})= (1,0,0)$, $(\mu_{2,1}, \mu_{2,2}, \mu_{2,3})= (0,1,0)$, $(\mu_{3,1}, \mu_{3,2}, \mu_{3,3})= (0,0,1)$.
	It is clear that if $s=1$, the optimal arm sequence is $(i,i,i)$ for any $i\in\{1,2,3\}$, and the unified oracle takes value $\sum^3_{t=1}\mu_{t,i}= 1$. 
    If $s=2$, then an optimal arm sequence is $(1,2,2)$ and the unified oracle equals to $\mu_{1,1}+\mu_{2,2}+ \mu_{3,2}= 2$. 
    If $s=3$, then the optimal arm sequence is $(1,2,3)$, and the unified oracle equals to $\mu_{1,1}+\mu_{2,2}+ \mu_{3,3}= 3$.
    Therefore, a larger switch budget generally leads to better performance of the oracle.
\end{remark}

To make the connection to dynamic and static oracles, note that when $s=1$,
we must have $a_1=\dots=a_T$. 
As a result, $\Oscr(1,T, \{\mu_{t,k}\}_{t\in[T],k\in[K]})$ is the performance of the best arm in hindsight, regardless of the nonstationarity of the mean rewards.
On the other hand, when $s=T$, then $O(T,T, \{\mu_{t,k}\}_{t\in[T],k\in[K]})$ is the performance of the dynamic oracle, because the switch budget is vaccuous and one can let $a_t = \argmax \{\mu_{t,k}, k\in [K]\}$.
As $s$ varies from $1$ to $T$, the unified oracle smoothly interpolates between static and dynamic oracles.

Next, we define the (pseudo) regret of a policy $\pi$,
given $T$, $s=S_T$, and $ \{\mu_{t,k}\}_{t\in[T],k\in[K]}$:
\begin{align*}
    R^{\pi} (T; S_T, \{\mu_{t,k} \}_{t\in[T],k\in[K]})
    &= \Oscr(S_T,T,\{\mu_{t,k} \}_{t\in[T],k\in[K]})
    - \E\left( \sum^T_{t=1} \mu_{t,A_t}\right)
\end{align*}
The expectation is taken with respect to the randomness of the history when implementing $\pi$ as well as the internal randomization of the policy itself.
As one may expect, the regret depends on how much and frequently the expected rewards $\mu_{t,k}$ change over time.
To control for that, we use a metric called the variational budget that has been introduced in the literature to track the cumulative variation of the mean rewards, denoted by $V_T$.
More precisely, consider the set of mean rewards
\begin{equation}
\begin{split}
    \cL(V)  = \Big \{ \{\mu_{t,k}\} \in [0,1]^{T \times K}: \sum_{t=1}^T\sup_{k\in [K]} |\mu_{t,k} - \mu_{t+1,k}|\leq V \Big \}
    \end{split}
\end{equation}
whose variation over the horizon is bounded by $V$.
See, e.g., \cite{besbes2015non,besbes2019optimal} for similar setups.
In the rest of the paper, we investigate the regret
\begin{equation}
    R^{\pi}(T; S_T,V_T)  = \sup_{\{\mu_{t,k}\} \in \cL(V_T)} R^{\pi}\Big (T;S_T,\left\{\mu_{t,k}\right\} \Big )
\end{equation}
as a function of $T$, the switch budget $S_T$ and the variational budget $V_T$.
Note that although both $S_T$ and $V_T$ are referred to as ``budget'', $S_T$ measures the conservativeness of the oracle while $V_T$ measures the nonstationarity of the environment.
\begin{remark}
When $S_T=1$ and $V_T= T$, the regret $R^{\pi}(T; S_T, V_T)$ reduces to the adversarial bandit problem benchmarked against a static oracle \citep*{auer1995gambling}.
	When $S_T= T$ and $V_T= o(T)$, the regret $R^{\pi}(T; S_T, V_T)$ reduces to the nonstationary bandit problem benchmarked against a dynamic oracle \citep*{besbes2015non}.
	
\end{remark}

Before introducing the algorithm and the analysis, we explicitly state what the initial knowledge of the agent.
The agent has the information of $K$, $T$, $S_T$ and $V_T$, as well as the fact that $\mu_{t,k},Y_{t,k}\in[0,1]$ for all $t$ and $k$.
The knowledge of $K$ is commonly assumed in the literature.
The knowledge of $T$ is mostly innocuous. 
If $T$ is unknown, we can apply the techniques in corollary 8.4 of \cite{auer2002nonstochastic} to modify the algorithm to achieve a similar result.
The switch budget $S_T$ is usually an intentional choice of the agent instead of a parameter in the problem: it encodes how conservative the agent is and the balance she wants to strike between dynamic and static oracles.
Therefore, it is reasonable to assume the knowledge.
Lastly, we assume the knowledge of the variational budget in our model for simplicity.
It is possible to extend to unknown $V_T$ using the techniques developed in \cite{cheung2019learning,auer2019adaptively}.
We leave it for future research.

\section{The Adaptive EXP3.S (AE3) Algorithm}\label{sec:algorithm}
In this section, we present an algorithm that achieves near-optimal regret benchmarked against the unified oracle.
The algorithm is referred to as the \emph{AE3} Algorithm. 
It is based on the EXP3.S algorithm (Algorithm~\ref{alg:exp3}) as a subroutine, which is proposed in \cite{auer2002nonstochastic}.
We briefly introduce the idea of the EXP3.S algorithm.
In Algorithm~\ref{alg:exp3}, the weights $w_{t,k}$ and the learning rate $\gamma$ determine the probability of choosing each arm at time $t$ (Step 3 and 4).
More precisely, with probability $\gamma$, the arms are chosen uniformly randomly (the second term in Step 3).
With probability $1-\gamma$, the arms are chosen with a probability proportional to the weights of the arms (the first term in Step 3).
Step 5 updates the weights of the arms based on the observed reward.
Note that $\hat X_{t, k}$ can be regarded as an unbiased estimator for $\mu_{t,k}$ and is used to update the weight exponentially.
EXP3.S is based on the well-known EXP3 algorithm \citep{auer1995gambling}, which is commonly used to handle adversarial bandit problems.
EXP3.S introduces another parameter $\alpha$ to update the weights, which adds some flexibility to scenarios where the number of switches in the optimal arm's identity is finite.

\begin{algorithm}[t]\caption{Exploration-Exploitation with Exponential weights (EXP3.S) }\label{alg:exp3}
\begin{algorithmic}[1]
\renewcommand{\algorithmicrequire}{\textbf{Input:}}
\REQUIRE{positive number $\alpha$, learning rate $\gamma$, arms $K$, horizon length $T$}
\STATE{Initilization: for any $k \in [K]$, set $w_{1,k} \gets 1$. }
\FOR{$t= 1,2,\cdots,T $}
\STATE{For each $k\in [K]$, set
$$
p_{t,k} = (1-\gamma) \frac{w_{t,k}}{\sum_{s=1}^K w_{t,s}} + \frac{\gamma}{K}.
$$}
\STATE Draw an arm $A_t$ from 
the probability distribution $\{ p_{t,k} \}_{k=1}^K$, receive a reward $Y_{t,A_t}$.
\STATE Set $\hat X_{t,A_t} = \frac{Y_{t,A_t}}{p_{t,A_t}}$, and set $\hat X_{t,k} = 0$  for $k \neq A_t$. Update
$$
w_{t+1,k} = w_{t,k} \exp \Big ( \frac{ \gamma \hat X_{t,k}}{K} \Big )+ \frac{e\alpha}{K} \sum^K_{k'=1} w_{t,k'}
, \text{ for all }k \in [K].
$$
\ENDFOR
\end{algorithmic}
\end{algorithm}

Note that the regret of the EXP3.S algorithm against the \emph{static oracle} has been studied in the literature.
For example, \cite{auer2002nonstochastic} shows that the regret is bounded by
$2\sqrt{TK\log(TK)}$ by choosing a proper learning rate $\gamma$.\footnote{We always use $\log$ for the natural logarithm in this paper.}
Furthermore, when the number of switches in the optimal arm's identity is bounded by $S$ (this corresponds to $S_T= S, V_T= T$ under our model setting), the regret of Algorithm~\ref{alg:exp3} is shown to be of order $\sqrt{TKS\log(TK)}$, by choosing a proper $\alpha$ and $\gamma$.

The main algorithm, Algorithm~\ref{alg:we3}, applies the EXP3.S algorithm with fine-tuned learning rates based on the problem instance, including $T$, $K$, $V_T$, and $S_T$.
In particular, if the switch budget $S_T$ is large (Step 1), then
we choose $\gamma$ to be of order $\cO((\frac{KV_T}{T})^{1/3})$.
On the other hand, if $S_T$ is relatively small (Step 4),
then we choose $\gamma$ to be $\cO((\frac{KS_T}{T})^{1/2})$.
One may think of $\gamma$ to depend on the minimum of $V_T$ and $S_T$ up to a polynomial order, which serves as the bottleneck of the oracle.

The AE3 algorithm operates based on the following intuitions.
When the value of $S_T$ is small, the unified oracle closely resembles a static oracle. In this case, the algorithm follows the adversarial framework described by \cite{auer2002nonstochastic} to achieve optimal learning rate tuning.
When the value of $S_T$ is large, the unified oracle becomes more similar to a dynamic oracle. Here, the variational budget $V_T$ becomes relevant, and the algorithm should adopt the nonstationary framework presented by \cite{besbes2019optimal} to tune the learning rate $\gamma$ while considering the variational budget.
The distinction in tuning the learning rate $\gamma$ arises from the nature of the oracle. When using a static oracle, the optimal $\gamma$ is of the order of $\sqrt{K/T}$.
However, with a stronger oracle, the optimal $\gamma$ should be determined differently.

A few remarks are in line. Firstly, when $S_T$ is large and the problem is closer to a nonstationary framework, \cite{besbes2019optimal}  propose another type of algorithm with optimal regret bounds, based on the idea of dividing the time periods into batches and restarting the EXP3 algorithm in each batch.
By sharing information within the same batch and ``forgetting'' information across different batches, this algorithm achieves near-optimal learning efficiency. 
In contrast, there is no need to restart the EXP3.S algorithm in Algorithm~\ref{alg:we3}, since it balances the tradeoff between information sharing and ``forgetting'' automatically through a careful joint tuning of parameters $\alpha$ and $\gamma$.
Secondly, 
Algorithm~\ref{alg:we3} can be adapted to the case of unknown time horizon $T$.
In fact, whenever $T$ is unknown, we can use Exp3.S as a subroutine over exponentially increasing pulls epochs $T_{\ell}= 2^\ell, \ell= 0,1,2,\cdots$, in a manner that is similar to the one described in corollary 8.4 in \cite{auer2002nonstochastic}, and obtain performance bounds of the same order. 

\begin{algorithm} 
	\renewcommand{\algorithmicrequire}{\textbf{Input:}}
	\renewcommand{\algorithmicensure}{\textbf{Output:}}
	\caption{AE3 Algorithm}
	\label{alg:we3}
	\begin{algorithmic}[1]
		\REQUIRE Arms $K$, horizon $T$, switch budget $S_T$, variational budget $V_T$
		\IF{$S_T\ge K^{-1/3} V_T^{2/3} T^{1/3}$} 
		\STATE Use EXP3.S algorithm with $\gamma= \min\{1, (\frac{4KV_T\log(KT)}{(e-1)^2T})^{1/3}\}$.
		\ENDIF 
		\IF{$S_T< K^{-1/3} V_T^{2/3} T^{1/3}$} 
		\STATE Use EXP3.S algorithm with  $\gamma= \min\left\{1, \sqrt{\frac{K(S_T\log (KT)+ e)}{(e-1)T}} \right\}$.
		\ENDIF 
	\end{algorithmic}  
\end{algorithm}

\subsection{Regret Analysis of Algorithm~\ref{alg:we3}}\label{sec:ub}
In this section, we provide an upper bound for the regret of Algorithm~\ref{alg:we3} with $\alpha= \frac{1}{T}$ and defer the detailed proof to the Appendix~\ref{app:proof_upper-bound}. 

\begin{theorem}\label{thm:ub}
Suppose $KV_T \leq T$, $K \geq 2$ and $S_T\log (KT)\ge e$.
The regret of Algorithm~\ref{alg:we3} with $\alpha= \frac{1}{T}$ is bounded by
\begin{equation*}
 R^{AE3}(T; S_T,V_T)\le \begin{cases}
 \bar{C} (K V_T \log(KT))^{1/3} T^{2/3}&   S_T> K^{-1/3} V_T^{2/3} T^{1/3}\\ 
4\sqrt{e-1} \sqrt{KS_TT\log(KT)} & S_T\le K^{-1/3} V_T^{2/3} T^{1/3},
\end{cases}
\end{equation*}
where $\bar{C}$ is an absolute constant.
\end{theorem}

There are a few implications of the results.
First, the two cases are separated by a threshold of $S_TV_T^{-2/3}$.
When the switch budget is large or the variational budget is small, the regret is independent of $S_T$ and is of the same order as that in \cite{besbes2015non,besbes2019optimal}.
This corresponds to the regime of nonstationary MAB.
When the switch budget is small or the variational budget is large, the regret is independent of $V_T$ and increases in $S_T$.
When $S_T= 1$, it is consistent with the regret in \cite{auer1995gambling},
corresponding to the regime of adversarial MAB.
The two regimes have distinct regret behaviors and thus display a phase transition.
Second, the threshold of $S_T$ that separates the two cases, satisfies
\(S_TV_T^{-2/3} = (T/K)^{1/3}\).
We may investigate a few extreme cases: if $V_T=\Omega(T)$ or $ S_T=O(T^{1/3})$, then the problem always falls into the adversarial regime and the regret is $\cO(\sqrt{S_TT})$.
The variational budget doesn't affect the regret anymore and the regret is solely determined by the switch budget.
Indeed at the boundary, the two regret expressions are of the same order except for a constant term and a term of order $\log (KT)$.


\section{Lower Bound}\label{sec:lb}
In this section, we derive a lower bound for the regret of any policy for the problem.
We show that the regret derived in Theorem~\ref{thm:ub} cannot be further improved.
\begin{theorem}\label{thm:lower-bound}
    For given $T$, $K$, $V_T$, $S_T$ and $K$ satisfying $T \geq K \geq 2$ and $V_T \leq T/K$, the regret for any policy $\pi$ is lower bounded by
    \begin{equation*}
R^{\pi}(T; S_T,V_T)  \geq
\begin{cases}
    \left (\frac{1}{16} -  \frac{ \sqrt{2  \log(4/3)}}{16}
    \right )  (KV_T)^{1/3} T^{2/3} &S_T \ge    K^{-1/3}  V_T^{2/3}T^{1/3}\\
  \left (\frac{1}{16} -  \frac{ \sqrt{2  \log(4/3)}}{16}
    \right )  \sqrt{  KS_TT}& S_T \le    K^{-1/3}  V_T^{2/3}T^{1/3}
\end{cases}
    \end{equation*}
\end{theorem}
We provide the detailed proof in the Appendix~\ref{app:proof_lower-bound}. 
We briefly discuss how we construct the problem instances under the two distinct scenarios for the lower bounds.
To do so, we group the time horizon into $m$ batches $\cT_1,\cdots, \cT_m$ of size $\Delta$ (possibly except the last batch $\cT_m$), where the value of $\Delta$ is specified later. We denote by $T_j$ the number of time periods within batch $\cT_j$.
It is easy to see that $T_j=\Delta$ except for $j=m$.

    We construct the following bandit instances:
    for periods in batch $i$ ($t\in \cT_i$), we uniformly randomly pick an arm $k_i \in [K]$ and set $\mu_{t,k_i} = 1/2 + \epsilon$ and $\mu_{t,s} = 1/2$ for $s \neq k_i$ in the entire batch.
    More precisely, we construct $K^m$ uniformly distributed bandit instances, denoted by $(k_1,\dots,k_m)\in [K]^m$.
    For an instance $(k_1,\dots,k_m)$, the mean reward satisfies
    \begin{equation}\label{eq:lb-mean}
        \mu_{t,k} =\begin{cases}
            1/2 + \epsilon & t\in \cT_i, k=k_i,\\
            1/2 & \text{otherwise}.\\
        \end{cases}
    \end{equation}
When $ S_T \ge    K^{-1/3}  V_T^{2/3}T^{1/3}$,
this case is similar to Theorem~1 of \cite{besbes2019optimal}.
We choose $\Delta= \lceil K^{1/3} (T/V_T)^{2/3}\rceil$ and $\epsilon = \min \Big \{ \frac{1}{4} \sqrt{  K/\Delta}, \frac{ V_T \Delta}{T} \Big \}$.
We set $\Delta\le T_m<2\Delta$.
It can be shown that $\epsilon \leq \frac{1}{4}$ so that the constructed instances satisfy $\mu_{t,k} \in [0,1]$. 
It is easy to check that the variational budget is satisfied.
Note that the optimal arm does not change within the same batch, so the optimal action sequence for the dynamic oracle does not have more than $m$ switches.
Because $m= \lfloor T/\Delta\rfloor \le S_T$, the unified oracle coincides with the dynamic oracle.
When $ S_T \le    K^{-1/3}  V_T^{2/3}T^{1/3} $,
we let $\Delta= \lceil T/S_T\rceil$ and $\epsilon = \min \Big \{ \frac{1}{4} \sqrt{  K/\Delta}, \frac{ V_T \Delta}{T} \Big \}$.
For batch $\cT_i$, we set
$\epsilon_i = \min \Big \{ \frac{1}{4} \sqrt{  K/W_i}, \frac{ V_T W_i}{2T} \Big \}$. Similar to the previous case, it can also be shown that $\epsilon \leq \frac{1}{4}$, $\mu_{t,k} \in [0,1]$, and the variational budget is satisfied.

By combining Theorems \ref{thm:ub} and \ref{thm:lower-bound}, we observe that the upper and lower bounds match each other in both scenarios, respectively, if we ignore the $\log (KT)$ factor and constant terms.

\section{Numerical Experiments}\label{sec:numerical}
In this section, we conduct numeric experiments to test the empirical performance of the AE3 algorithm.
For simplicity, we suppress the subscript of $V_T$ and $S_T$.
Guided by Theorems~\ref{thm:ub} and~\ref{thm:lower-bound}, we consider two regimes:
(i) $ S \ge    K^{-1/3}  V^{2/3}T^{1/3}$, i.e., the switch budget $S$ is relatively large; (ii) $ S \le    K^{-1/3}  V^{2/3}T^{1/3}$, i.e., $S$ is relatively small.
We report the experimental details in the rest of this section.

We first consider the regime  $ S \ge    K^{-1/3}  V^{2/3}T^{1/3}$ and conduct three experiments to understand the performance of  our AE3 algorithm against the variational budget $V$, the number of arms $K$, and the number of periods $T$, respectively.
In each experiment, given the parameters $K$, $V$, $S$ and $T$, we generate problem instances by using the construction of lower bounds for $ S \ge K^{-1/3}  V^{2/3}T^{1/3}$ following Theorem \ref{thm:lower-bound}.
Specifically, we calculate $\Delta =\lceil K^{1/3} (T/V)^{2/3}\rceil$ and group the time periods along horizon $T$ into different batches of size $\Delta$ (except the last batch). 
Within batch $\cT_k$, we uniformly pick one arm $i_k$ and set its expected reward to be $\mu_{t,i_k} = \frac{1}{2}+ \epsilon$ where $\epsilon  = \min \{ \frac{1}{4} \sqrt{  \frac{K}{ \Delta}},  \frac{V \Delta}{T}\}$, and set the expected rewards of the remaining arms to be $\mu_{t,s} = \frac{1}{2}$ for all $s \neq i_k$. Under such construction, the expected rewards of all arms remain unchanged within each epoch, and vary between batches.

\begin{figure}[t]
\begin{center}
\includegraphics[width=0.4\linewidth]{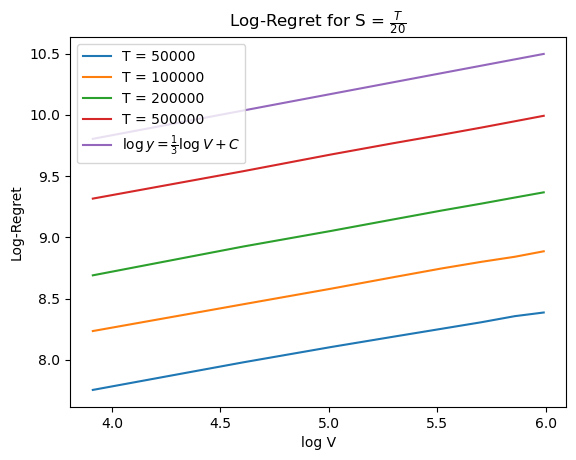}
\includegraphics[width=0.4\linewidth]{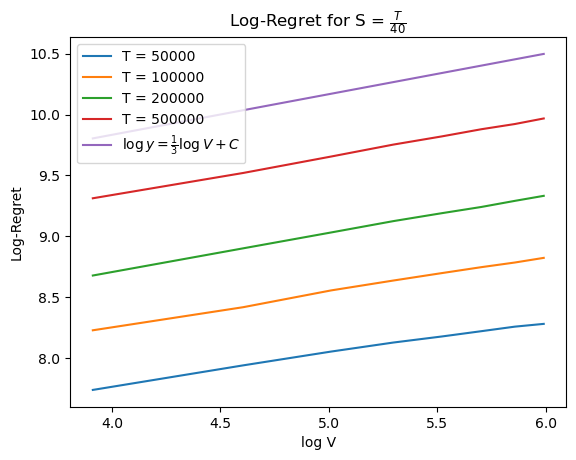}
\caption{ Empirical Log-Regret when $S\ge K^{-1/3}  V^{2/3}T^{1/3}$ for $S \in \{ T/20,T/40\}$, $K = 10$, $V \in [50, 400]$}
\label{fig:01}
\end{center}
\end{figure}

For the regret versus $V$, we set $K = 10$, $V \in [50,400]$,  $T \in \{ 50000,100000,200000,500000 \} $, and $S  \in \{ \frac{T}{20} ,\frac{T}{40} \}$.
For each instance, we run 10 independent instances of the AE3 algorithm, and compute the empirical regret by taking their average. 
To understand the performance of our algorithm against the variation budget $V$, we plot $\log V$ against the log-regret for different choices of $T$. We  provide one additional line of slope $1/3$ for benchmark comparison. We summarize the results in Figure \ref{fig:01}.

\begin{figure}[t]
\begin{center}
\includegraphics[width=0.4\linewidth]{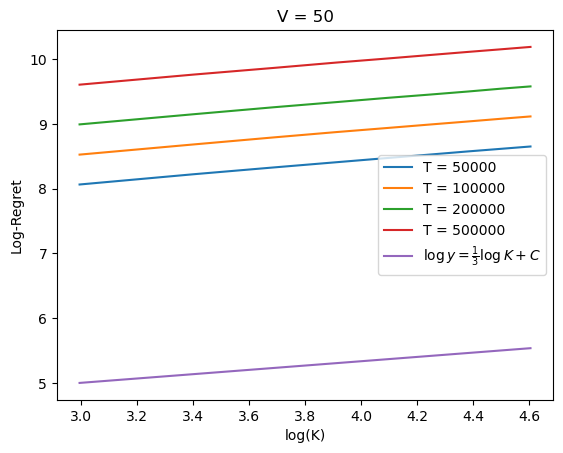}
\includegraphics[width=0.4\linewidth]{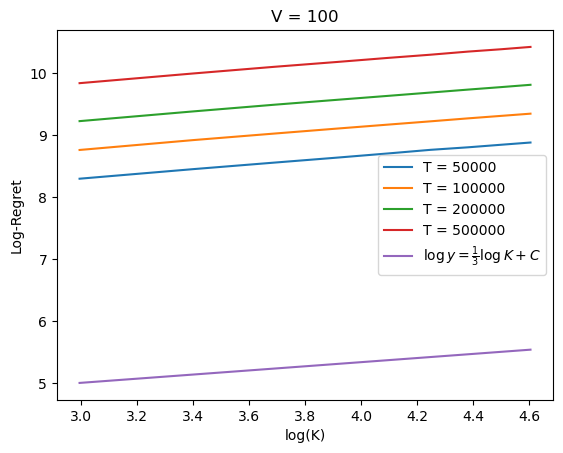}
\caption{ Empirical Log-Regret when $S\ge K^{-1/3}  V^{2/3}T^{1/3}$ for $V = 50$, $S=T/10$, $V \in \{ 50,100\}$, $K \in [20,100]$  }
\label{fig:02}
\end{center}
\end{figure}

For the regret versus $K$, we set $V \in \{ 50,100\}$, $S= \frac{T}{10}$, $K \in [20,100]$, and test the AE3 algorithm over $T \in \{ 50000,100000,200000,500000 \}$. We compute the empirical regrets by averaging over 10 independent replications, and plot $\log K$ against log-regret in Figure \ref{fig:02}. For benchmark comparison, we also provide one additional line of slope $1/3$.

\begin{figure}
\begin{center}
\includegraphics[width=0.45\linewidth]{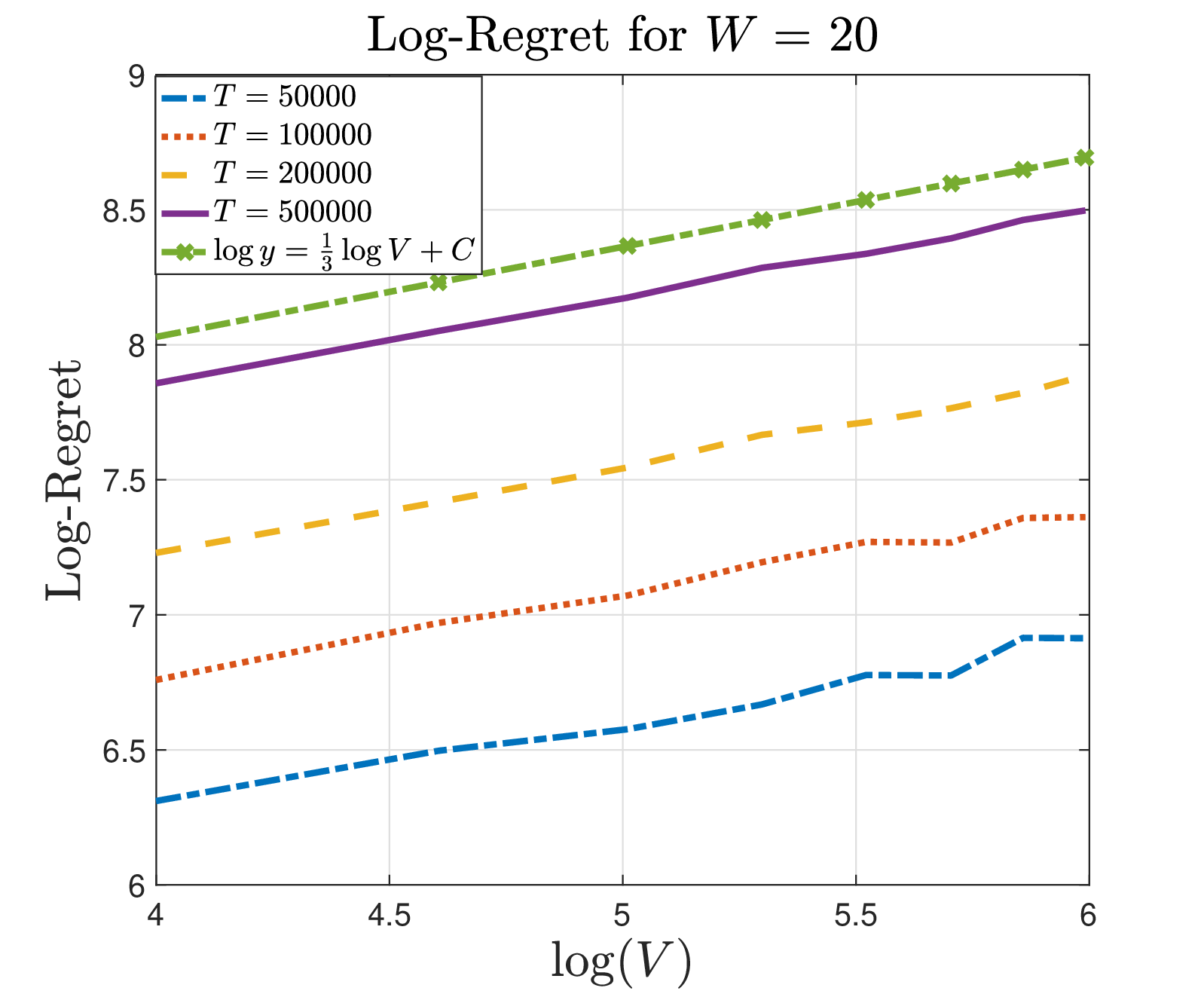}
\includegraphics[width=0.45\linewidth]{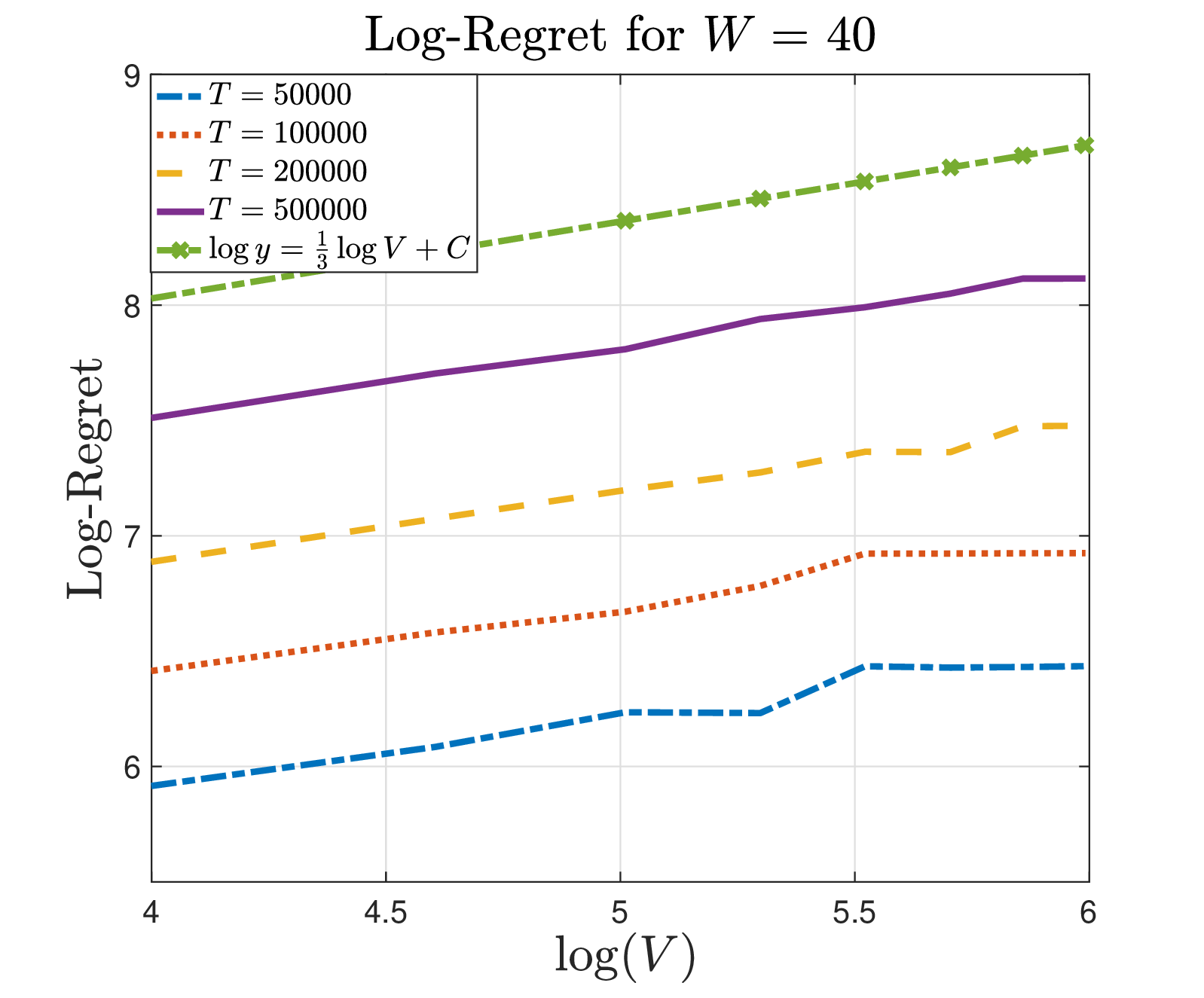}
\caption{ Empirical Log-Regret when $S\ge K^{-1/3}  V^{2/3}T^{1/3}$ for $S \in \{ T/20, T/40\}$ , $K= 10$, $T \in [50000,500000]$}
\label{fig:03}
\end{center}
\end{figure}

For the regret versus $T$, we set $K=10$, $S  \in \{ \frac{T}{20} ,\frac{T}{40} \}$, $T \in [50000,500000]$, and $V \in \{ 50,100,150,200 \} $. Similar to the previous experiments, we plot $\log T$ against the log-regret, provide one additional line of slope $2/3$, and summarize the results in Figure \ref{fig:03}.

In Figures \ref{fig:01}, \ref{fig:02} and \ref{fig:03}, we observe that the slopes of $\log V$, $\log K$, and $\log T$ against the log-regret are close to $1/3$, $1/3$, and $2/3$, respectively, matching our theoretical rates in Theorem \ref{thm:ub} that the expected regret incurred by AE3 algorithm is bounded by $\tilde \cO((KV)^{1/3}T^{2/3})$ when $ S \ge    K^{-1/3}  V^{2/3}T^{1/3}$.

%

\begin{figure}[t]
\begin{center}
\includegraphics[width=0.4\linewidth]{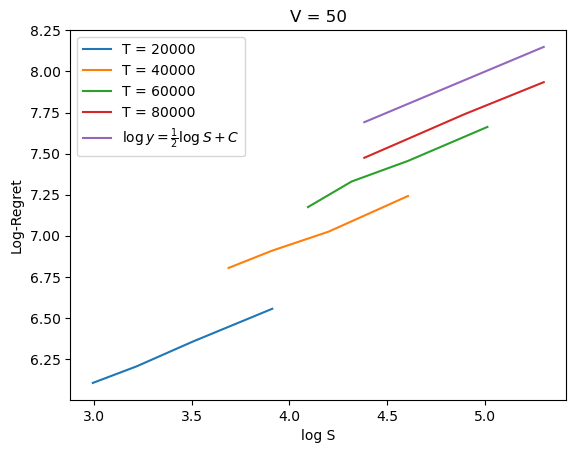}
\includegraphics[width=0.4\linewidth]{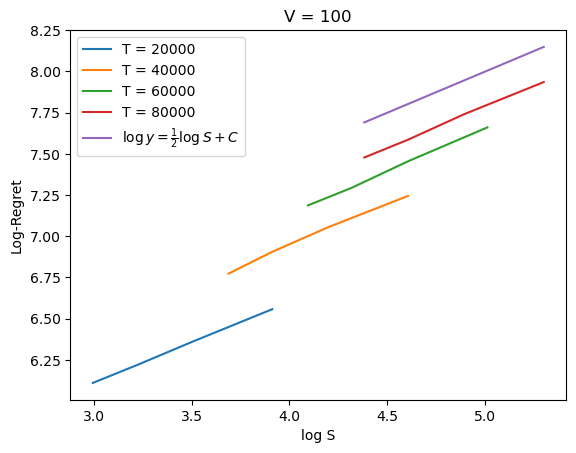}
\caption{ Empirical Log-Regret when $S< K^{-1/3}  V^{2/3}T^{1/3}$ for $V \in \{50$,100\}, $K = 10$, $W \in [400,1000]$ }
\label{fig:04}
\end{center}
\end{figure}

We then consider the regime  $ S \le    K^{-1/3}  V^{2/3}T^{1/3}$. 
Similarly, we calculate $\Delta =\lceil T/S \rceil$ and group the time periods along horizon $T$ into different batches of size $\Delta$ (except the last batch).
In this case, we conduct two sets of experiments to study the performance of the AE3 algorithm against the switch budget $S$ and number of arms $K$, respectively. For regret versus $S$, we set $V \in \{ 50, 100\}$, $K = 10$, $S \in [T/1000,T/400]$,
and generate problem instances by using the construction of lower bounds when $ S \le    K^{-1/3}  V^{2/3}T^{1/3}$ in the proof of Theorem \ref{thm:lower-bound}, with $T \in \{ 20000,40000,60000,80000 \}$. For each instance, we run 10 independent instances of our AE3 algorithm, and compute the empirical regret by taking their average. 
We plot $\log V$ against the log-regret for different choices of $T$, and provide one additional one of slope $-1/2$ for benchmark comparison. We summarize the results in Figure \ref{fig:04}.

\begin{figure}[t]
\begin{center}
\includegraphics[width=0.4\linewidth]{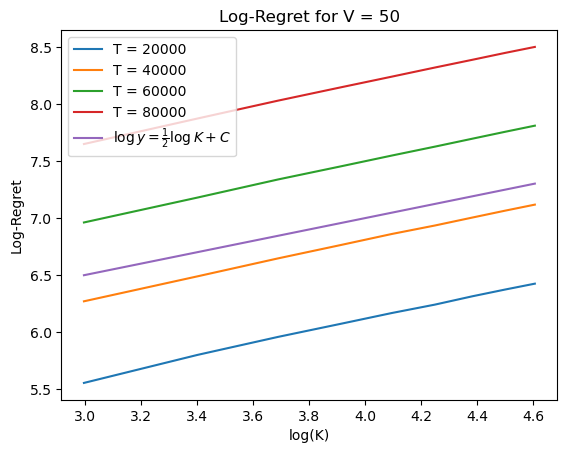}
\includegraphics[width=0.4\linewidth]{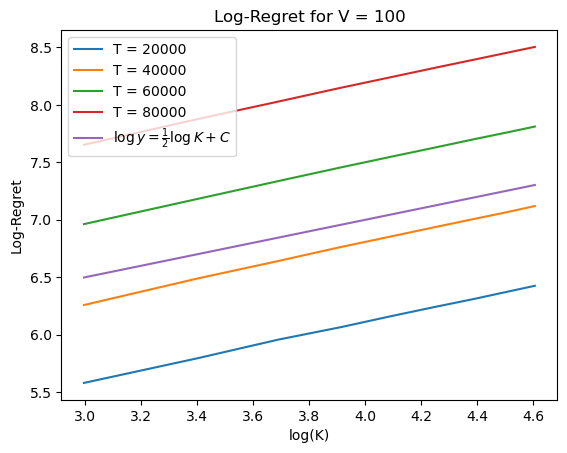}
\caption{ Empirical Log-Regret when $S< K^{-1/3}  V^{2/3}T^{1/3}$ for $V \in \{ 50,100\}$, $S = T/400$, $K \in [20,100]$ }
\label{fig:05}
\end{center}
\end{figure}

For regret versus $K$, we generated problem instances with $W =400$ , $K\in [20,100]$, $T \in [5000,40000]$, $S  \in \{ \frac{T}{50} ,\frac{T}{100} \}$. We plot $\log K$ against the log-regret averaged over 10 independent simulations, provide one additional line of slope $1/2$, and summarize the results in Figure \ref{fig:05}.

From Figures \ref{fig:04} and \ref{fig:05}, we observe that the slopes of $\log S$ and $\log K$ against log-regret are close to $-1/2$ and $1/2$, respectively, matching our theoretical claim in Theorem \ref{sec:lb} that the regret incurred by our AE3 algorithm is bounded by $\cO( \sqrt{KST })$.

Furthermore, from Figures \ref{fig:02} and \ref{fig:05}, we observe that the number of arms $K$ exhibits distinct impacts on the regret in the two cases. In particular, the expected regret incurred by AE3 algorithm is upper bounded by $\cO(K^{1/3})$ when $S$ is relatively large, and is bounded by $\cO(K^{1/2})$ when $S$ is relatively small. This also matches the upper bounds provided by Theorem \ref{thm:ub} under the two cases.

%
%
%
%
%

\section{Future Directions}\label{sec:future-direction}
This paper provides a general formulation that smoothly bridges the nonstationary and adversarial bandit problems studied in the literature.
The proposed unified oracle allows the agent to fine-tune the expectation and conservativeness of the algorithm.
We believe that it provides a flexible framework to handle nonstationary environments.
However, this study also points to several important open directions.
In particular, does there exist a policy or an algorithm that can always achieve the optimal regret while being agnostic to the choice of switch budget $S_T$?
Since $S_T$ only affects the performance of the oracle and does not affect the objective performance of the policy (expected total reward), one would expect that the policy may not need the input of $S$.
This is not the case of our AE3 algorithm and it remains an open direction.
It is worth mentioning that relatedly \cite{auer2019adaptively} present a challenge in developing an algorithm that achieves optimal regret bounds in an adversarial setting without requiring pre-defined adjustments based on the number of arm changes.



%
\bibliographystyle{chicago}
\bibliography{ref.bib}





\begin{APPENDIX}{}

\section{Proofs of Upper Bounds}\label{app:proof_upper-bound}
\textit{Proof of Theorem \ref{thm:ub}:}
Firstly, we study the regret when $S_T <  K^{-1/3}V_T^{2/3} T^{1/3} $. 
The idea of proof mainly follows \cite{auer2002nonstochastic}.
We denote by $W_t$ the sum of all weights at time $t$: $W_t= \sum^K_{k=1} w^k_t$, and ${\bf j}^*= (j^*_1,\cdots,j^*_T)\in \argmax\limits_{H(j_1,\cdots,j_t)\le S} \sum^T_{t=1}\mu_{t,j_t}$ be a sequence of actions with $H({\bf j}^*)= S_T$ (the case of $H({\bf j}^*)< S_T$ can be proven similarly).
Thus, $(1,2\cdots,T)$ is partitioned into segments
\begin{align*}
	[T_1,\cdots,T_2), [T_2,\cdots, T_3),\cdots, [T_S,\cdots, T_{S+1})
\end{align*}
where $T_1=1, T_{S+1}= T+1$ and $j^*_{T_s}=\cdots= j^*_{T_{s+1}-1}$ for each segment $s\in \{1,\cdots, S\}$.
Following the proof of Theorem 8.1 in \cite{auer2002nonstochastic}, we have 
\begin{align*}
	\frac{W_{t+1}}{W_t}
	&\le 1+ \frac{\gamma/K}{1-\gamma} X_t^\pi + \frac{(e-2)(\gamma/K)^2}{1-\gamma}\sum^K_{k=1}\hat{X}^k_t + e\alpha,
\end{align*}
where $X^\pi_t$ is the random reward obtained under the policy $\pi$ at time $t$.
Taking logarithms on both sides of and summing over all $t\in [T_s, T_{s+1})$, we get
\begin{align}
	\log \frac{W_{T_{s+1}}}{W_{T_s}}
	&\le \frac{\gamma/K}{1-\gamma} \sum^{T_{s+1}-1}_{t= T_s} X_t^\pi + \frac{(e-2)(\gamma/K)^2}{1-\gamma}\sum^{T_{s+1}-1}_{t= T_s}\sum^K_{k=1} \hat{X}^k_t + e\alpha (T_{s+1}- T_s).
	\label{eq:tech-0}
\end{align}
Let $j$ be the action such that $j^*_{T_s}=\cdots= j^*_{T_{s+1}-1}= j$.
Since
\begin{align*}
	w^j_{T_{s+1}}
	\ge w^j_{T_{s}+1}\exp\left\{ \frac{\gamma}{K}\sum^{T_{s+1}-1}_{t= T_s+ 1}\hat{X}^j_t \right\}
	\ge \frac{e\alpha}{K}W_{T_s} \exp\left\{ \frac{\gamma}{K}\sum^{T_{s+1}-1}_{t= T_s+ 1}\hat{X}^j_t \right\}
	\ge \frac{\alpha}{K}W_{T_s} \exp\left\{ \frac{\gamma}{K}\sum^{T_{s+1}-1}_{t= T_s}\hat{X}^j_t \right\}
\end{align*}
where the last inequality holds because $\gamma \hat{X}^j_t/K \le 1$.
We have
\begin{align}
	\label{eq:tech-1}
	\log \frac{W_{T_{s+1}}}{W_{T_s}}
	\ge \log \frac{w^j_{T_{s+1}}}{W_{T_s}}
	\ge \log (\frac{\alpha}{K}) + \frac{\gamma}{K}\sum^{T_{s+1}-1}_{t= T_s}\hat{X}^j_t. 
\end{align}
Putting together \eqref{eq:tech-0} and \eqref{eq:tech-1}, we have
\begin{align*}
	\sum^{T_{s+1}-1}_{t= T_s} X_t^\pi
	\ge (1-\gamma) \sum^{T_{s+1}-1}_{t= T_s}\hat{X}^j_t- \frac{K\log (K/\alpha)}{\gamma} - (e-2)\frac{\gamma}{K} \sum^{T_{s+1}-1}_{t= T_s}\sum^K_{k=1} \hat{X}^k_t- \frac{e\alpha K}{\gamma}(T_{s+1}- T_s).
\end{align*}
Summing over all segments $s = 1,\cdots,S_T$, taking expectation with respect to the
noisy rewards and actions of algorithm Exp3.S, and using the fact that $\mu^k_t\le 1$ for all $k, t$, we have the following result
\begin{align*}
	R^{\pi}(T; S_T,V_T)
	\le \frac{KS_T\log (K/\alpha) + Ke\alpha T}{\gamma} + (e-1) \gamma T.
\end{align*}
By choosing $\alpha= 1/T$ and $\gamma= \min\left\{1, \sqrt{\frac{K(S\log (KT)+ e)}{(e-1)T}} \right\}$, the desired results follow.

Next we look at the regret when $S_T \ge   K^{-1/3}V_T^{2/3} T^{1/3} $.
Since $a_t^*= \max_{k\in [K]}\mu_{t,k}$ is defined as the best arm for time period $t$, we have $\sum_{t=1}^T \max_{k\in [K]}\mu_{t,k}\ge \max\limits_{H(j_1,\cdots,j_t)\le S_T} \sum^T_{t=1}\mu_{t,j_t}= \Oscr(S_T,T,\{\mu_{t,k} \}_{t\in[T],k\in[K]})$.
Therefore, the regret is bounded by that against the dynamic oracle for any policy $\pi$, i.e.,
\begin{equation}\label{eq:case_one_upper}
 R^{\pi}(T; S_T,V_T)  \leq  \sum_{t=1}^T \max_{k\in [K]}\mu_{t,k} - \sum_{t=1}^T \EE[ \mu_{t,A_t}].
\end{equation}
Instead of establishing an upper bound of $R^{\pi}(T;S,V)$, we only need to provide an upper bound for the regret with respect to the dynamic oracle \eqref{eq:case_one_upper}.
The proof is completed by Theorem 3 of \cite{besbes2019optimal}.

In summary, we obtain the following worst-case performance of Algorithm~\ref{alg:we3}:
When $S_T> K^{-1/3} V_T^{2/3} T^{1/3}$, the regret is bounded by $ R^{AE3}(T; S_T,V_T)
 \leq \bar{C} (K V_T \log(KT))^{1/3} T^{2/3}$
for some absolute constant $\bar{C}$.
When $S_T\le K^{-1/3} V_T^{2/3} T^{1/3}$, the regret is bounded by
$R^{AE3}(T; S_T,V_T)\le    4\sqrt{e-1} \sqrt{KS_TT\log(KT)}$.
\qed 


\section{Proofs of Lower Bounds} \label{app:proof_lower-bound}
\textit{Proof of Theorem~\ref{thm:lower-bound}:}
    To construct problem instances to establish lower bounds, we first group the time horizon $[1,T]$ into $m= \lfloor \frac{T}{\Delta}\rfloor$ batches $\cT_1,\cdots, \cT_m$ of size $\Delta$ (perhaps except $\cT_m$), where
    $\cT_j= \{t: (j-1)\Delta+ 1\le t\le \max\{T, j\Delta\}\}$ for all $j=1,\cdots, m$.
	Let $T_j$ be the number of time periods in $\cT_j$, it is obvious that $T_j= \Delta$ for $j=1,\cdots,m-1$, and $\Delta\le T_m< 2\Delta$.

    We construct the following bandit instances:
    for periods in batch $i$ ($t\in \cT_i$), we uniformly randomly pick an arm $k_i \in [K]$ and set $\mu_{t,k_i} = 1/2 + \epsilon_i$ and $\mu_{t,s} = 1/2$ for $s \neq k_i$ in the entire batch.
    More precisely, we construct $K^m$ uniformly distributed bandit instances, denoted by $(k_1,\dots,k_m)\in [K]^m$.
    For an instance $(k_1,\dots,k_m)$, the mean reward satisfies
    \begin{equation}\label{eq:lb-mean}
        \mu_{t,k} =\begin{cases}
            1/2 + \epsilon_i & t\in \cT_i, k=k_i,\\
            1/2 & \text{otherwise}.\\
        \end{cases}
    \end{equation}
The choice of $\epsilon_i$ will be specified shortly.
The random reward has a Bernoulli distribution based on the mean in \eqref{eq:lb-mean}.
We establish the lower bound for the following cases.

Case one: $S_T \ge   K^{-1/3} V_T^{2/3}T^{1/3}$. 
This case is similar to Theorem~1 of \cite{besbes2019optimal}.
We set $\Delta =\lceil K^{1/3} (T/V_T)^{2/3}\rceil$ and  $\epsilon = \min\left \{ \frac{1}{4} \sqrt{  \frac{K}{\Delta}},  \frac{V_T \Delta}{T}\right \}$. 
For each batch $\cT_j$, since $V_T\le T/K$, we have
\begin{equation*}
\sqrt{\frac{  K}{\Delta}} \leq \sqrt{\frac{ K}{K ^{1/3} (T/V_T)^{2/3}}}  = \sqrt{\frac{(KV)^{2/3}}{T^{2/3}}} \leq  1.
\end{equation*}
Therefore, it is easy to see that
\begin{equation}\label{eq:epsilon}
\epsilon_j = \min \left\{ \frac{1}{4} \sqrt{\frac{  K}{ \Delta }}, \frac{V_T \Delta }{T}\right\} \le  \frac{1}{4},
\end{equation}
and the constructed instances \eqref{eq:lb-mean} satisfy $\mu_{t,k}\in [0,1]$.
Moreover, because $\mu_{t,k}$ remains constant for $t$ in a batch, the total variation is bounded by
\begin{equation}\label{eq:variation-budget-lb}
\sum_{t=1}^{T-1} \sup_{k \in [K]} |\mu_{t,k} - \mu_{t+1,k}|
\leq \sum_{j=1}^{m-1}  \epsilon
\leq m\epsilon\leq
m  \frac{V_T \Delta}{T} \le V_T,
\end{equation}
where we have used the fact that when $t$ moves from batch $\cT_{j}$ to batch $\cT_{j+1}$, the change in $\mu_{t,k}$ is at most $\epsilon$ for all $k$.
Therefore, the variational budget is satisfied.

Next we consider the regret for the MAB instances constructed above.
For any policy $\pi$, the regret $R^{\pi}(T;S, (k_1,\dots,k_m))$ under the instance $(k_1,\dots,k_m)$
can be expressed as
\begin{align*}
    R^{\pi}(T;S_T, (k_1,\dots,k_m))& = \max\limits_{H(j_1,\cdots,j_t)\le S_T} \sum^T_{t=1}\mu_{t,j_t}-\sum_{t=1}^T\E[\mu_{t,\pi_t}]\\
                                 & = \sum_{i=1}^m \left( \frac{1}{2}+\epsilon\right)\Delta - \sum_{i=1}^m\sum_{t\in \cT_i}\left(\frac{1}{2}+\epsilon \ind\{\pi_t= k_i\}\right)\\
                                 & =\epsilon\sum_{i=1}^m  \sum_{t\in \cT_i} \ind\{\pi_t \neq k_i\}.
\end{align*}
The second equality follows from the fact that $S_T \ge  \lfloor T/\Delta\rfloor= m$ and the construction of the instances \eqref{eq:lb-mean}.
We next use a variant of the Assouad method \cite{yu1997assouad} to show the lower bound.
More precisely, because the instances satisfy the variational budget, we have
\begin{align}
    R^\pi(T;S_T, V_T)&\ge \max_{(k_1,\dots,k_m)\in [K]^m}R^{\pi}(T;S_T, (k_1,\dots,k_m))\ge \frac{1}{K^m}\sum_{(k_1,\dots,k_m)\in [K]^m}R^{\pi}(T;S_T, (k_1,\dots,k_m))\notag\\
    &\ge \frac{\epsilon }{K^m}\sum_{(k_1,\dots,k_m)\in [K]^m}\sum_{i=1}^{m}  \sum_{t\in \cT_i}  \PP(\pi_t\neq k_i)\notag\\
    &\ge \frac{\epsilon }{K^m}\sum_{i=1}^{m} \sum_{ (k_1,\dots,k_{i-1},k_{i+1},\dots,k_m)\in [K]^{m-1} }\sum_{k_i\in [K]} \sum_{t\in \cT_i} \PP(\pi_t\neq k_i)\notag\\
    &\ge \frac{\epsilon }{K^m}\sum_{i=1}^{m} \sum_{ (k_1,\dots,k_{i-1},k_{i+1},\dots,k_m)\in [K]^{m-1} } \left(KT_i-\sum_{k_i\in [K]}\sum_{t\in \cT_i} \PP(\pi_t=k_i)\right).\label{eq:regret-lb-analysis1}
\end{align}
Next we focus on the term $\sum_{k_i\in [K]}\sum_{t\in \cT_i} \PP(\pi_t= k_i)$.
Denote the probability measure of $\left\{\pi_t\right\}_{t\in \cT_i}$ generated by the instance $(k_1,\dots,k_{i-1}, k_i)$ by $ \PP_k$ when $k_i=k\in [K]$.
Here we omit the dependence on $(k_1,\dots,k_{i-1})$ which is treated as given in analyzing the term.
Our goal is thus to provide an upper bound for $\sum_{k\in [K]}\sum_{t\in \cT_i} \PP_k(\pi_t= k)$.
Applying Lemma A.1 from \cite{auer2002nonstochastic}, we have
\begin{align*}
    \sum_{t\in \cT_i}\PP_k(\pi_t= k)\le \frac{1}{K} \sum_{t\in \cT_i}\sum_{k'\in [K]}\PP_{k'}(\pi_t=k) + \frac{T_i}{2}\sqrt{- \frac{1}{K} \sum_{t\in \cT_i}\sum_{k'\in [K]}\PP_{k'}(\pi_t=k) \log(1-4\epsilon_i^2)}.
\end{align*}
Therefore, we have
\begin{align*}
    \sum_{k\in [K]}\sum_{t\in \cT_i} \PP_k(\pi_t= k_i) & \le\frac{1}{K}\sum_{t\in \cT_i} \sum_{k'\in [K]}\sum_{k\in [K]}\PP_{k'}(\pi_t=k) \\
    &\quad + \frac{T_i}{2}\sum_{k\in [K]}\sqrt{- \frac{1}{K} \sum_{t\in \cT_i}\sum_{k'\in [K]}\PP_{k'}(\pi_t=k) \log(1-4\epsilon^2)}\\
    &\le T_i+ \frac{T_i}{2}\sqrt{- \sum_{t\in \cT_i}\sum_{k'\in [K]}\sum_{k\in [K]}\PP_{k'}(\pi_t=k) \log(1-4\epsilon^2)}\\
    &\le T_i+ \frac{T_i}{2}\sqrt{- KT_i \log(1-4\epsilon^2)}.
\end{align*}
In the second inequality, we have used Jensen's inequality and $\sum_{k\in[K]}\PP_{k'}(\pi_t=k)=1$.
Therefore, we have
\begin{align}
    \eqref{eq:regret-lb-analysis1}&\ge \frac{\epsilon }{K^m}\sum_{i=1}^{m} \sum_{ (k_1,\dots,k_{i-1},k_{i+1},\dots,k_m)\in [K]^{m-1} }\left((K-1)T_i-\frac{T_i}{2}\sqrt{- KT_i \log(1-4\epsilon^2)}\right)\notag\\
                                  &= \frac{\epsilon }{K}\sum_{i=1}^m \left((K-1)T_i-\frac{T_i}{2}\sqrt{- KT_i \log(1-4\epsilon^2)}\right).\label{eq:regret-lb-analysis2}
\end{align}
Because $- \log(1-x) \leq 4\log(4/3) x$ for $x \in [0,1/4]$, setting $x = 4\epsilon^2$ yields $ -\log(1-4\epsilon^2) \leq 16 \log(4/3) \epsilon^2$.
Therefore, we can further bound \eqref{eq:regret-lb-analysis2} as
\begin{align*}
    \eqref{eq:regret-lb-analysis2} &\ge
    \sum_{i=1}^m \epsilon T_i  \left (
    \frac{K-1}{K} - \frac{2 \epsilon}{K} \sqrt{ KT_i \log(4/3)}
    \right )
    \\
    &\ge
    \sum_{i=1}^m \epsilon T_i  \left (
    \frac{1}{2} - 2 \epsilon \sqrt{ \frac{  T_i \log(4/3)} {K} }
    \right ).
\end{align*}
In the last inequality, we have used the fact that $K\ge 2$.
Recall that $\epsilon = \min \{ \frac{1}{4} \sqrt{  \frac{K}{\Delta}},  \frac{V_T \Delta}{T}\}$.
Therefore, we have
\begin{equation}\label{eq:lower_1}
\begin{split}
R(T; S_T,V_T) & \geq   \sum_{i=1}^m \epsilon T_i  \left (
    \frac{1}{2} - 2 \epsilon \sqrt{ \frac{  T_i \log(4/3)} {K} }
    \right )
    \geq  \sum_{i=1}^m \epsilon T_i  \left (
    \frac{1}{2} -  \frac{ \sqrt{ 2 \log(4/3)}}{2}
    \right ) \\
&    \geq   \min \Big  \{ \frac{1}{4}T \sqrt{ \frac{K}{\Delta}},  V_T\Delta \Big \} \cdot  \left (
    \frac{1}{2} -  \frac{ \sqrt{ 2 \log(4/3)}}{2}
    \right ).
\end{split}
\end{equation}
Because $\Delta =\lceil K^{1/3}(T/V_T)^{2/3} \rceil\ge K^{1/3}(T/V_T)^{2/3}$, we have
$T \sqrt{ \frac{K}{\Delta}}\le V_T\Delta$, we further obtain
\begin{equation}\label{eq:lower_2}
\begin{split}
R(T; S_T,V_T)
&    \geq  \left (
    \frac{1}{8} -  \frac{ \sqrt{2 \log(4/3)}}{8}
    \right )  T \sqrt{ \frac{K}{\Delta}}.
\end{split}
\end{equation}
Finally, by noting that $V_T \le T/K$, we have
\begin{equation*}
\Delta\le K^{1/3}(T/V_T)^{2/3}+ 1\le 2 K^{1/3}(T/V_T)^{2/3}.
\end{equation*}
By substituting the above inequality into \eqref{eq:lower_2}, we conclude that
\begin{equation*}
\begin{split}
R(T; S,V)
&    \geq  \left (
    \frac{1}{16} -  \frac{ \sqrt{2  \log(4/3)}}{16}
    \right )  (KV_T)^{1/3} T^{2/3}.
\end{split}
\end{equation*}
This completes the proof for case one.

Case two: $S_T <   K^{-1/3} V_T^{2/3}T^{1/3}$. 
In this case, we set $\Delta= \lceil\frac{T}{S_T}\rceil $ and 
$\epsilon = \min \Big \{ \frac{1}{4} \sqrt{  K/\Delta}, \frac{ V_T \Delta}{T} \Big \}$.
Then we have
$$
\epsilon \leq \frac{1}{4}\sqrt{  \frac{K}{\Delta}}\le \frac{1}{4}\sqrt{  \frac{KS_T}{T}} < \frac{1}{4} \left( \frac{KV_T}{T}\right)^{1/3} \leq \frac{1}{4},
$$
where the last inequality uses the fact that  $V_T \leq T/K$.
Similar to \eqref{eq:variation-budget-lb} in case one, the variational budget constraint is satisfied.
By following the same steps as \eqref{eq:lower_1}, we have
\begin{equation*}
R(T; S_T,V_T) \geq  \min \Big  \{ \frac{1}{4}T \sqrt{ \frac{K}{\Delta}},  V_T\Delta \Big \} \cdot  \left (
    \frac{1}{2} -  \frac{ \sqrt{ 2 \log(4/3)}}{2}
    \right ) 
\end{equation*}
Since $S_T <   K^{-1/3} V_T^{2/3}T^{1/3}$, we have $\Delta\ge T/S_T\ge K^{1/3} V_T^{-2/3} T^{2/3}$, therefore $T \sqrt{ \frac{K}{\Delta}}\le V_T\Delta$, and \eqref{eq:lower_2} still holds.

Finally, by noting that $\Delta\le T/S_T + 1\le 2T/S_T$, we have
\begin{equation*}
R(T; S_T,V_T)\geq \left ( \frac{1}{16} -  \frac{ \sqrt{2  \log(4/3)}}{16}
    \right )  T\sqrt{KS_T/T}=  \left (\frac{1}{16} -  \frac{ \sqrt{2  \log(4/3)}}{16}
    \right ) \sqrt{KS_TT}.
\end{equation*}
This completes the proof for Case two.
\qed 
\end{APPENDIX}

\end{document}